\documentclass[conference]{IEEEtran}
\IEEEoverridecommandlockouts

\usepackage{cite}
\usepackage{amsmath,amssymb,amsfonts}
\usepackage{algorithmic}
\usepackage{graphicx}
\usepackage{textcomp}
\usepackage{xcolor}
\usepackage{siunitx}
\usepackage{booktabs}
\usepackage{multirow}
\usepackage{graphicx}
\usepackage{bbding}
\usepackage[pagebackref,breaklinks,colorlinks]{hyperref}
\usepackage{comment}
\usepackage[capitalize]{cleveref}
\crefname{section}{Sec.}{Secs.}
\Crefname{section}{Section}{Sections}
\Crefname{table}{Table}{Tables}
\crefname{table}{Tab.}{Tabs.}

\begin{document}


\title{TH\"{O}R-MAGNI \emph{Act}: Actions for Human Motion Modeling in Robot-Shared Industrial Spaces\\

\thanks{This work was supported by the Wallenberg AI, Autonomous Systems and Software Program (WASP) and by the EU Horizon 2020 No. 101017274 (DARKO).}
}


\author{\IEEEauthorblockN{Tiago Rodrigues de Almeida}
\IEEEauthorblockA{\textit{AASS}, 
\textit{\"{O}rebro University}\\
\"{O}rebro, Sweden \\
tiago.almeida@oru.se}
\and
\IEEEauthorblockN{Tim Schreiter}
\IEEEauthorblockA{\textit{PercInS}, 
\textit{Technical University of Munich}\\
Munich, Germany \\
tim.schreiter@tum.de}
\and
\IEEEauthorblockN{Andrey Rudenko}
\IEEEauthorblockA{\textit{ Corporate Research},
\textit{Robert Bosch GmbH}\\
Stuttgart, Germany \\
andrey.rudenko@de.bosch.com}
\and
\IEEEauthorblockN{Luigi Palmieri}
\IEEEauthorblockA{\textit{ Corporate Research},
\textit{Robert Bosch GmbH}\\
Stuttgart, Germany \\
luigi.palmieri@de.bosch.com}
\and
\IEEEauthorblockN{Johannes A. Stork}
\IEEEauthorblockA{\textit{AASS}, 
\textit{\"{O}rebro University}\\
\"{O}rebro, Sweden \\
johannesandreas.stork@oru.se}
\and
\IEEEauthorblockN{Achim J. Lilienthal}
\IEEEauthorblockA{\textit{PercInS}, 
\textit{Technical University of Munich}\\
Munich, Germany \\
achim.j.lilienthal@tum.de}
}

\maketitle

\begin{abstract} 


Accurate human activity and trajectory prediction are crucial for ensuring safe and reliable human-robot interactions in 
dynamic environments, such as industrial settings, with mobile robots.
Datasets with fine-grained action labels for moving people in industrial environments with mobile robots are scarce, as most existing datasets focus on social navigation in public spaces.
This paper introduces the TH\"{O}R-MAGNI~\emph{Act} dataset, a substantial extension of the TH\"{O}R-MAGNI dataset,
which captures participant movements alongside robots in diverse semantic and spatial contexts.
TH\"{O}R-MAGNI~\emph{Act} provides 8.3 hours of manually labeled participant actions derived from egocentric 
videos recorded via eye-tracking glasses. 
These actions, aligned with the provided TH\"{O}R-MAGNI motion cues, follow a long-tailed 
distribution with diversified acceleration, velocity, and navigation distance profiles. 
We demonstrate the utility of TH\"{O}R-MAGNI \emph{Act} for two tasks: action-conditioned trajectory 
prediction and joint action and trajectory prediction. 
We propose two efficient transformer-based models that outperform the baselines to address these tasks.
These results underscore the potential of TH\"{O}R-MAGNI \emph{Act} 
to develop predictive models for enhanced human-robot interaction in complex environments.

\end{abstract}

\section{Introduction} \label{sec:study_overview}







Human movement and actions are shaped by many factors that collectively define the context in which individuals navigate and interact
\cite{rudenko20}. 
These factors may be internal, such as the person's own tasks, goals, and preferences, as well as external, i.e., coming from the environment, such as the location of obstacles, affordances, and semantically meaningful regions~\cite{cao2020long}. Robots can detect external factors and infer certain internal factors, using them to anticipate future motion and activities~\cite{li2014prediction,zhu2023cliff,gorlo2024} to predictively assist and navigate safely and efficiently~\cite{mavrogiannis2023core}.
Especially in industrial environments, robots face complex yet structured activities of people and interactions with other workers and robots~\cite{benmessabih2024online}. Datasets containing accurate labels of human motion in such environments are rare, with most works focusing on social navigation in public spaces where the dominant activity is walking and standing~\cite{Malla_2020_CVPR,ehsanpour22,Girase_2021_ICCV,Rasouli_2019_ICCV}.

%
%

%
The recent TH\"{O}R-MAGNI dataset aims to address this gap by providing a large-scale indoor motion capture recording of human navigation and robot interaction~\cite{schreiter2024th}.
It encompasses five distinct scenarios that simulate typical activities in industrial environments, such as transportation of various objects, interaction with robots, and goal-oriented navigation alone and in groups. TH\"{O}R-MAGNI includes 3.5 hours of motion from 40 participants and 8.3 hours of egocentric videos from 16 participants over five recording days. The activity labels in TH\"{O}R-MAGNI are, however, limited to the invariable role of each participant (referred to as agent class), representing a complex activity assigned to the participant for the duration of the experiment. These roles have been shown to improve trajectory prediction~\cite{almeida24_ral} and describe gaze behavior~\cite{schreiter2024gaze}, but fine-grained action labels would be required to model the separate sub-tasks and their durations in each activity.
%
%
%
%
%

\begin{figure}[t]
    \centering
    \includegraphics[width=\linewidth]{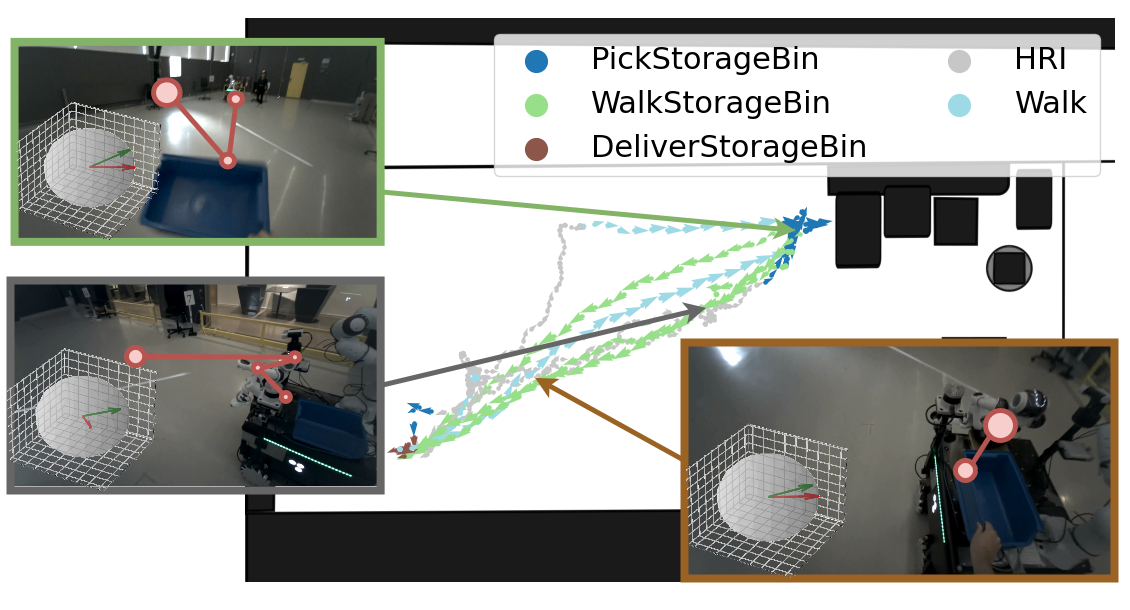}
    
    \vspace{-3.5mm}
    \caption{Action annotations for a 4-minute recording of a person carrying storage bins while interacting with a mobile robot, synchronized with the motion capture data. Inset images display snapshots from gaze overlaid videos, featuring visualizations of head orientation vector (\textbf{red}) and gaze vector (\textbf{green}). The length of the arrows on the map denotes the velocity magnitude.}
    \vspace{-4mm}
    \label{fig:intro}
\end{figure}

In this paper, we present TH\"{O}R-MAGNI~\emph{Act} -- an extension of the original dataset, which provides 8.3 hours of fine-grained actions derived from the first-person view videos of participants wearing eye-tracking glasses. Our TH\"{O}R-MAGNI~\emph{Act} is unique in aligning action labels with high-quality multi-modal first-person gaze and third-person motion capture data, as shown in Fig.~\ref{fig:intro}. These labels enable the robot to anticipate long-term human motion trajectories and actions.

To demonstrate the utility of TH\"{O}R-MAGNI~\emph{Act}, we present two transformer-based trajectory prediction frameworks: (1) role- and action-conditioned trajectory prediction, extending~\cite{almeida24_ral}, and (2) joint prediction of future trajectory and corresponding actions via multi-task learning~\cite{Zhang_22}. Our results show how incorporating action labels can improve the performance of these predictive models.
TH\"{O}R-MAGNI \emph{Act} and the corresponding scripts are stored in a publicly accessible repository\footnote{\url{https://github.com/tmralmeida/thor-magni-actions}}. Documentation on how to use and visualize the dataset can be found in the same repository.

\section{TH\"{O}R-MAGNI Act} \label{sec:thor_magni_act}

\subsection{Experimental Design}

The TH\"{O}R-MAGNI data acquisition, conducted in a laboratory, simulates industrial logistics settings to explore diverse human-human and human-robot interactions~\cite{schreiter2024th}. 
The dataset comprises five scenarios, distinguished by the spatial layout, the mode of the robot operation, and specific tasks assigned to the participants. 
These tasks or agent classes represent the high-level activities assigned to the person for the duration of a 4-minute recording session, such as: \emph{Carrier--Box}, \emph{Carrier--Bucket}, \emph{Carrier--Large Object}, \emph{Visitors--Alone}, \emph{Visitors--Group}, and \emph{Visitors--Alone HRI}. 
In each session, one to three participants wear eye-tracking glasses that capture egocentric video data. In~\cite{schreiter2024th}, we provided standardized instructions to ensure natural behavior, informing participants that the experiment aimed to evaluate the robot's perception of human actions. 
In particular, participants were asked to carry stacks of boxes and buckets between the source and target locations (\emph{Carrier--Box} and \emph{Bucket}). Some participants moved a large poster stand (\emph{Large Object}) in pairs of two. \emph{Visitors} navigated freely between goal points, drawing a random card to determine their next destination. 
\emph{Visitor--Alone HRI} involved both passive interactions (passing the robot in close proximity) and active interactions (joint navigation to goal points) with a mobile robot. The \emph{Carrier--Storage Bin HRI} collaboratively with the robot transported storage bins between random goal points.

\subsection{Action Annotations} \label{subsec:action_annotation}

\begin{figure}[t]
    \centering
    \vspace{-2.5mm}
    \includegraphics[width=\linewidth]{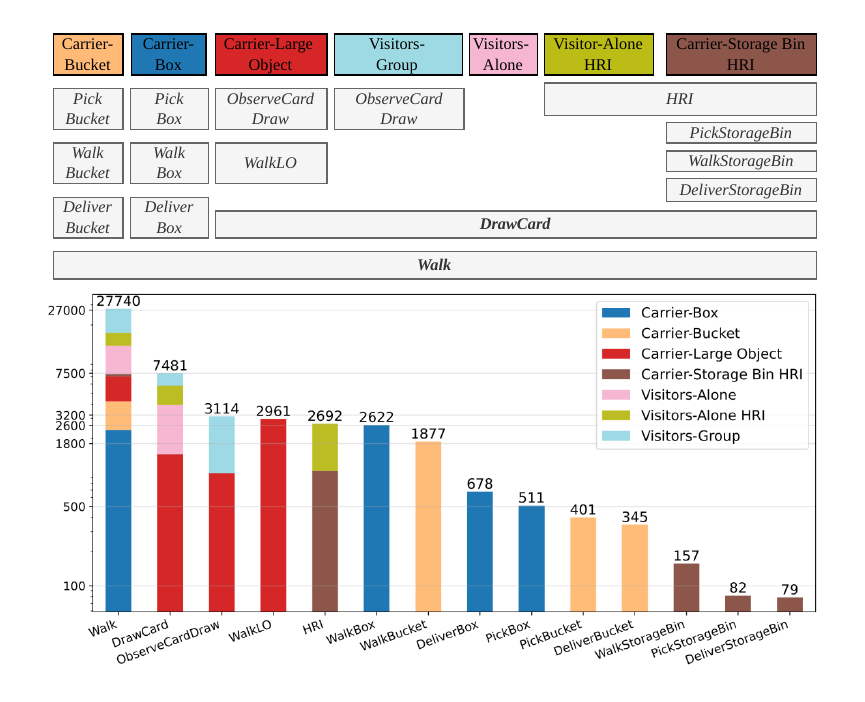}
    \vspace{-11mm}
    \caption{\textbf{Top:} Agent class-actions mapping. Grey boxes denote actions, colored boxes represent the agent classes. \textbf{Bottom:} Distribution of action classes in \emph{log-scale} sorted by descending order, with colors indicating agent classes.}
    \vspace{-4mm}
    \label{fig:action-class}
\end{figure}

In this work, we define a set $\mathcal{A}$ of 14 unique action labels: {\em Walk}, {\em DrawCard}, observing another person drawing a card at a goal point ({\em ObserveCardDraw}), moving a larger object ({\em WalkLO}), 
{\em PickBucket}, {\em WalkBucket}, {\em DeliverBucket}, {\em PickBox}, {\em WalkBox}, {\em DeliverBox}, 
{\em PickStorageBin}, {\em WalkStorageBin}, {\em DeliverStorageBin}, and {\em HRI}, based on the existing agent classes in TH\"{O}R-MAGNI. 
Each agent class is associated with specific actions, while some actions are shared across
different agent classes (see \cref{fig:action-class} for an overview). 
An agent class is constant in all trajectories of a particular agent, whereas an action class may change at every time step.
Consequently, this data extension provides finer labeling of internal factors (e.g., goal-driven actions) that can influence human motion. 
In particular, the action classes {\em Walk} and {\em DrawCard} are shared across multiple agent classes, 
indicating that trajectories involving these actions are likely to have similar characteristics, 
even when performed by agents of different agent classes.

We annotated these actions in the entire TH\"{O}R-MAGNI dataset using the 8.3 hours of egocentric videos and the ``Event Marker'' feature in the eye-tracking software~\cite{TobiiProLabManual}.
Markers were placed at initial fixations indicating action switches, such as reaching for objects or bending to deliver items. For ambiguous switches where hands were not visible, we selected subsequent suitable fixations during the activity. To enhance accuracy for those ambiguous switches, we leverage additional cues from eye-tracker data, including IMU and audio information. 
This process was curated by hand, ensuring high annotation quality.

\subsection{Dataset Statistics}

TH\"{O}R-MAGNI \emph{Act} statistics are computed for non-overlapping 8-second trajectory segments, 
in line with common trajectory prediction benchmarks~\cite{Kothari2020HumanTF}.
\cref{fig:action-class} bottom presents the distribution of action classes in log-scale, along with their representation across different agent classes. 
Although the dataset's action classes follow a long-tailed distribution, it includes novel action labels specific to human tasks 
and mobile robot interactions, setting it apart from existing social navigation datasets. 
Along with motion cues and gaze vectors, these labels 
support research on egocentric action prediction models from visual input and gaze pattern analysis.

\begin{figure}[t]
    \centering
    \vspace{-7mm}
    \includegraphics[width=\linewidth]{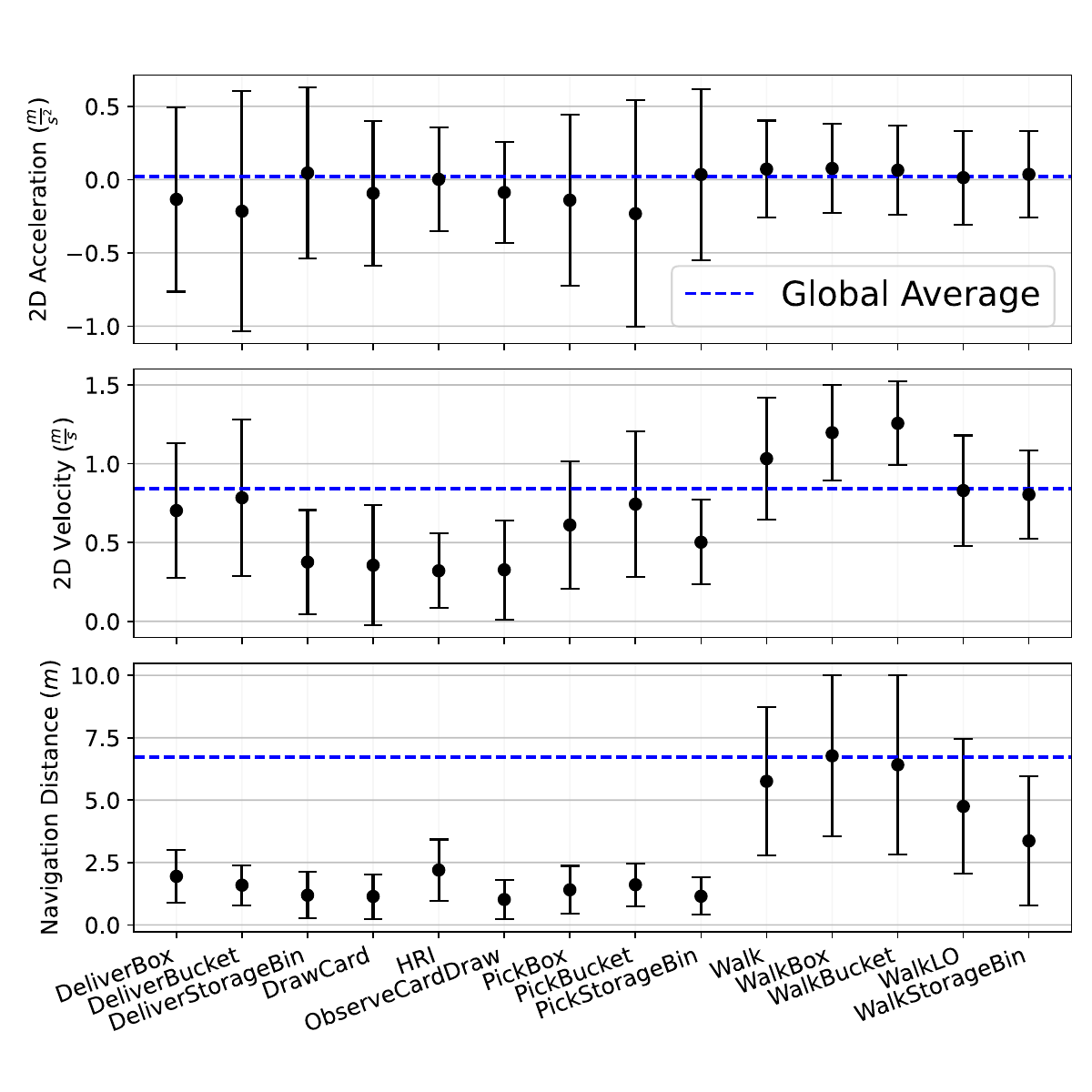}
    \vspace{-12mm}
    \caption{\textbf{Top}: 2D acceleration (mean $\pm$ one standard deviation), where values near zero indicate constant velocity.
    \textbf{Middle}: 2D velocity (mean $\pm$ one standard deviation), where values near zero correspond to static actions.
    \textbf{Bottom}: navigation distance (mean $\pm$ one standard deviation), where values near zero indicate static actions and higher values reflect walking actions.}
    \vspace{-4mm}
    \label{fig:stats}
\end{figure}

\cref{fig:stats} presents the average and standard deviation of acceleration, velocity, and navigation distance of motion in each action class, along with the corresponding global metrics (aggregated across all 8-second segments).
For acceleration, static actions such as picking up or delivering an object result in small negative accelerations, while walking actions generally show constant velocities or small positive accelerations. 
Consequently, in terms of velocity, static actions fall below the global average, whereas actions like {\em WalkBox} and {\em WalkBucket} involve higher velocities compared to {\em WalkLO} or {\em WalkStorageBin}, where participants move alongside the robot.
Finally, distance correlates with the acceleration and velocity trends, highlighting distinct action classes and further demonstrating the diversity and complexity of the dataset.

\section{Motion and Action Prediction Methods}\label{sec:usage}

In this section, we introduce two examples of tasks this dataset can be used for: \emph{action-conditioned 
trajectory prediction} (TP) and \emph{multi-task learning for joint trajectory and action prediction} (MTL), along with  
the respective proposed models.
We segment the 8-second trajectories, referred to as \emph{tracklets}, into observed and prediction parts, adhering to previous trajectory prediction benchmarks~\cite{Kothari2020HumanTF}. 
The observed horizon spans \SI{3.2}{s} (8 time steps), while the prediction horizon extends to \SI{4.8}{s} (12 time steps).
The observed tracklets are denoted as $\mathbf{S} = (\mathbf{s}_t)^{O}_{t=1}, O=8$, where the states $\mathbf{s}_t$ comprise 2D positions, 
velocities, and the corresponding action class $a$, represented as $\mathbf{s}_t = (x, y, \dot{x}, \dot{y}, a)$. 
The \emph{future} of an observed tracklet $\mathbf{Y_S}$ consists of 2D velocities, $\mathbf{Y_S} = ((\dot{x_t}, \dot{y_t}))^{T_P}_{t=O+1}$ of length $L = 12$, which are subsequently converted into future positions $\mathbf{P_S}$.
The future sequence of actions temporally aligned with $\mathbf{Y_S}$ is denoted by $\mathbf{a_S} = (a_t)_{t=O+1}^{T_P}$, $a_t\in \mathcal{A}$.

\subsection{Action-conditioned Trajectory Prediction} \label{subsec:act_cond}

The goal is to predict the future of a tracklet conditioned on the observed actions and agent class, 
$\psi_\mathrm{TP} \colon (\mathbf{S}_k, C_k) \mapsto \mathbf{Y}_{{\mathbf{S}}_k}$,
where $\mathbf{Y}_{\mathbf{S}_k}$ is the future corresponding to the observed tracklet $\mathbf{S}_k$.
The training data for this task, $\{(\mathbf{S}_k, C_k, \mathbf{P}_{\mathbf{S}_k})\}_k$, consists of triplets of observed tracklets, ground truth agent class labels, 
and ground truth future positions.

The $\psi_\mathrm{TP}$ model has an encoder-decoder structure as in~\cite{Kothari2020HumanTF,almeida24_ral}. The encoder $E$ consists of an embedding mapping (a single-hidden layer multilayer perceptron or MLP) followed by a transformer-based encoder~\cite{vaswani17}. 
The encoded features are then concatenated with the agent class embeddings and processed through the decoder network $D_\mathrm{T}$ (a two-hidden layer MLP) to generate the future sequence of velocities, $\mathbf{Y_S}$.
\cref{fig:mtl_methods}, excluding the yellow branch, depicts the graphical representation of $\psi_\mathrm{TP}$. 
The blue dotted arrow in the figure indicates the baseline model, which operates without agent class conditioning or action classes in $\mathbf{S}$.
We train $\psi_\mathrm{TP}$ with the Mean Squared Error (MSE) loss:
\begin{equation}
      L_\mathrm{TP}(\mathbf{P_S}, \mathbf{\hat{P}_S}) = \frac{1}{L} \sum_{j=O+1}^{T_P} \lVert \mathbf{p}^j -\mathbf{\hat{p}}^j  \rVert^2_2,
\label{eq:l_traj}
\end{equation}
where $\mathbf{p}^{j} = (x, y)$ is the ground truth 2D position at time step $j$ and $\mathbf{\hat{p}}^j$ is the corresponding 
prediction.

\subsection{Multi-Task Learning for Trajectory and Action Prediction}

The goal is to predict the future of a tracklet and the corresponding sequence of actions,
$\psi_\mathrm{MTL} \colon (\mathbf{S}_k, C_k) \mapsto (\mathbf{Y}_{\mathbf{S}_k}, \mathbf{a}_{\mathbf{S}_k})$,
where $\mathbf{Y}_{\mathbf{S}_k}$ is the future tracklet and $\mathbf{a}_{\mathbf{S}_k}$ the future sequence of actions 
corresponding to the observed tracklet $\mathbf{S}_k$.
The training data for this task, $\{(\mathbf{S}_k, C_k, \mathbf{P}_{\mathbf{S}_k}, \mathbf{a}_{\mathbf{S}_k})\}_k$, consists of quadruples of observed tracklets, ground truth agent class labels, 
ground truth future positions and sequence of actions. 

\begin{figure}
    \centering
    \vspace{-3mm}
    \includegraphics[width=\linewidth]{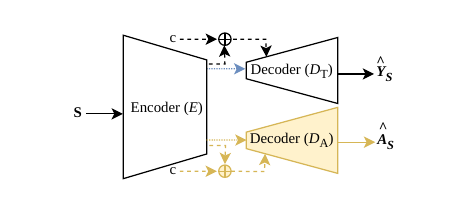}
    \vspace{-10mm}
    \caption{Action-conditioned models and multi-task learning methods (additional \textbf{yellow} branch).
    \textbf{Dashed} arrows indicate methods using agent class, while \textbf{dotted} arrows represent baseline models where $\mathbf{S}$ excludes actions in the trajectory prediction task.}
    \vspace{-4mm}
    \label{fig:mtl_methods}
\end{figure}

The $\psi_\mathrm{MTL}$ model is similar to the previously described $\psi_\mathrm{TP}$ model (see \cref{subsec:act_cond}). 
The key difference is the additional decoder, $D_\mathrm{A}$, which shares the same network configuration as $D_\mathrm{T}$. 
This decoder generates probabilities for the sequence of actions at each future time step, denoted as $\mathbf{A_S}^{L\times N_\mathrm{A}}$, where $N_\mathrm{A}=|\mathcal{A}|$.
The final actions $\mathbf{a}_{\mathbf{S}}$ are determined by applying the \emph{argmax} operator to these probabilities.
\cref{fig:mtl_methods}, including the yellow branch, depicts the graphical representation of $\psi_\mathrm{MTL}$, whose baseline consists of two models tailored to each task.
We train $\psi_\mathrm{MTL}$ using a weighted loss function that combines trajectory prediction (as defined in \cref{eq:l_traj}) and sequence of actions prediction, the latter with cross-entropy loss:
\begin{equation} 
  L_\mathrm{MTL}(\mathbf{P_S}, \mathbf{\hat{P}_S}, \mathbf{a_S}, \mathbf{\hat{a}_S}) = L_\mathrm{TP} (\mathbf{P_S}, \mathbf{\hat{P}_S}) + \lambda L_\mathrm{A} (\mathbf{a_S}, \mathbf{\hat{a}_S}), 
  \label{eq:l_mtl} 
\end{equation} 
where $\lambda$ is a weighting factor for the action class prediction term, and $L_\mathrm{A} (\mathbf{a_S}, \mathbf{\hat{a}_S}) = -\frac{1}{L} \sum_{j=O+1}^{T_P} \sum_{m=1}^{N_\mathrm{A}} a^j_{m} \log(\hat{a}^j_{m})$, where $\hat{a}^j_{m}$ is the predicted probability for class $m$ at time step $j$, and $a^j_m$ is a binary indicator of the ground truth for class $m$ at time step $j$.
In our experiments, we tested $\lambda=1$.

\section{Experiments and Results} \label{sec:experiments}

This section presents the data, evaluation setup, and results -- both qualitative and quantitative -- demonstrating the use of action labels alongside motion cues (i.e., positions, velocities, and agent classes) from TH\"{O}R-MAGNI~\emph{Act}.

\textbf{Target Scenarios.}
For our analysis, we merged Scenarios 2  (static robot) and 3 (moving robot) data from TH\"{O}R-MAGNI, as these scenarios encompass a more diverse set of agent classes and, consequently, a broader vocabulary of action classes.
These scenarios comprise a total of 5 agent classes: \emph{Carrier--Box}, \emph{Carrier--Bucket}, \emph{Carrier--Large Object}, \emph{Visitors--Alone}, and \emph{Visitors--Group}. 
In addition, they include 10 action classes: {\em DrawCard}, {\em Walk}, {\em WalkLO}, {\em PickBucket}, {\em WalkBucket}, {\em DeliverBucket}, {\em ObserveCardDraw}, {\em PickBox}, {\em WalkBox}, and {\em DeliverBox}. 
In total, the dataset contains 1227 trajectories.
%

\textbf{Evaluation Setup.} 
To evaluate the proposed models, we employ 5-fold cross-validation.
In the prediction results, we use \emph{Average} and \emph{Final Displacement Errors} (ADE and FDE in meters). 
ADE quantifies the average $\ell_2$ distance between the ground truth and predicted trajectory. 
FDE measures the $\ell_2$ distance between the final predicted position and corresponding ground truth.
In the action prediction results, we use accuracy (ACC) and F1 score (F1), both $\in [0,1]$. 
Accuracy represents the proportion of correct action predictions relative to the total number of instances. 
F1 score calculates the harmonic mean of precision and recall, providing a more balanced measure of the model's performance.
To ensure robust evaluation, we compute all metrics' mean and standard deviation across the validation folds.

\textbf{Quantitative Results.} \cref{tab:tp_results} shows the results for the action-conditioned trajectory prediction task (TP) with 
various cues settings.
It demonstrates that actions (third row) are powerful cues for trajectory prediction, outperforming the baseline (first row), with further improvements when combined with agent class information (last row). 
The additional parameters result from the action classes included in the input layer.

\begin{table}[t]
  \centering
  \caption{Action-conditioned trajectory prediction results, with bold values indicating superior performance of our agent and action class-aware models compared to the baseline.}
  \label{tab:tp_results}
  \vspace{-1mm}
  \resizebox{0.8\columnwidth}{!}{%
  \begin{tabular}{@{}c|cc|c|c@{}}
  \toprule
  \textbf{Model} &
    \textbf{\begin{tabular}[c]{@{}c@{}}Agent \\ Class\end{tabular}} &
    \textbf{\begin{tabular}[c]{@{}c@{}}Actions\\ Class\end{tabular}} &
    \textbf{\begin{tabular}[c]{@{}c@{}}ADE \\ FDE\end{tabular}} &
    \textbf{\begin{tabular}[c]{@{}c@{}}Number\\ Parameters (K)\end{tabular}} \\ \midrule
  BASELINE &
     &
     &
    \begin{tabular}[c]{@{}c@{}}0.71$\pm$0.03\\1.37$\pm$0.05\end{tabular} &
    36.7 \\ \specialrule{.2em}{.1em}{.1em}
  \multirow{6}{*}{OURS} &
  \Checkmark &
     &
    \begin{tabular}[c]{@{}c@{}}0.68$\pm$0.03\\1.30$\pm$0.07\end{tabular} &
    38.1 \\ \cmidrule{2-5}
   &
     &
     \Checkmark &
    \begin{tabular}[c]{@{}c@{}}0.69$\pm$0.03\\1.31$\pm$0.07\end{tabular} &
    37.3 \\ \cmidrule{2-5}
   &
   \Checkmark &
   \Checkmark &
    \textbf{\begin{tabular}[c]{@{}c@{}}0.67$\pm$0.03\\1.28$\pm$0.07\end{tabular}} &
    38.7 \\ \bottomrule
  \end{tabular}%
  }
  \vspace{-3mm}
  \end{table}

\cref{tab:mtl_results} shows the results for the joint action and trajectory prediction task (MTL).
We show the best baseline for action prediction where $\mathbf{s}_t = (x, y, \dot{x}, \dot{y}, a)$.
The results show that observed action sequences are crucial for strong performance in action prediction (second row versus third and fourth rows).
The best MTL approach can perform strongly in trajectory and action prediction simultaneously, outperforming baselines in trajectory prediction and matching single-task models in action prediction (last row).
Our MTL method (46.8K) is also more efficient than the baselines (36.7K+42.6K).

\begin{table}[t]
  \centering
  \caption{Comparative multi-task learning results, with bold values showing superior performance.}
  \vspace{-1mm}
  \label{tab:mtl_results}
  \resizebox{\columnwidth}{!}{%
  \begin{tabular}{@{}c|c|c|c|c|c@{}}
  \toprule
  \textbf{Model} &
    \textbf{\begin{tabular}[c]{@{}c@{}}Agent \\ Class\end{tabular}} &
    \textbf{\begin{tabular}[c]{@{}c@{}}Actions\\ Class\end{tabular}} &
    \textbf{\begin{tabular}[c]{@{}c@{}}ADE\\ FDE\end{tabular}} &
    \textbf{\begin{tabular}[c]{@{}c@{}}ACC\\ F1\end{tabular}} &
    \textbf{\begin{tabular}[c]{@{}c@{}}Number\\ Parameters (K)\end{tabular}} \\ \midrule
  BASELINES &
     &
     &
    \begin{tabular}[c]{@{}c@{}}0.71$\pm$0.03\\ 1.37$\pm$0.05\end{tabular} &
    \textbf{\begin{tabular}[c]{@{}c@{}}0.85$\pm$0.01\\ 0.85$\pm$0.01\end{tabular}} &
    36.7+42.6 \\ \specialrule{.2em}{.1em}{.1em}
  \multirow{8}{*}{OURS} 
   & \Checkmark
     &
     &
    \textbf{\begin{tabular}[c]{@{}c@{}}0.68$\pm$0.04\\ 1.29$\pm$0.08\end{tabular}} &
    \begin{tabular}[c]{@{}c@{}}0.62$\pm$0.02\\ 0.61$\pm$0.02\end{tabular} &
    46.3 \\ \cmidrule(l){2-6} 
   &
     & \Checkmark
     &
    \begin{tabular}[c]{@{}c@{}}0.70$\pm$0.03\\ 1.33$\pm$0.07\end{tabular} &
    \begin{tabular}[c]{@{}c@{}}0.83$\pm$0.01\\ 0.83$\pm$0.01\end{tabular} &
    43.3 \\ \cmidrule(l){2-6} 
   & \Checkmark
     & \Checkmark
     &
    \begin{tabular}[c]{@{}c@{}}0.70$\pm$0.04\\ 1.32$\pm$0.08\end{tabular} &
    \textbf{\begin{tabular}[c]{@{}c@{}}0.85$\pm$0.01\\ 0.85$\pm$0.01\end{tabular}} &
    46.8 \\ \bottomrule
  \end{tabular}%
  }
  \vspace{-1mm}
  \end{table}

\begin{figure}[t]
  \includegraphics[width=\linewidth]{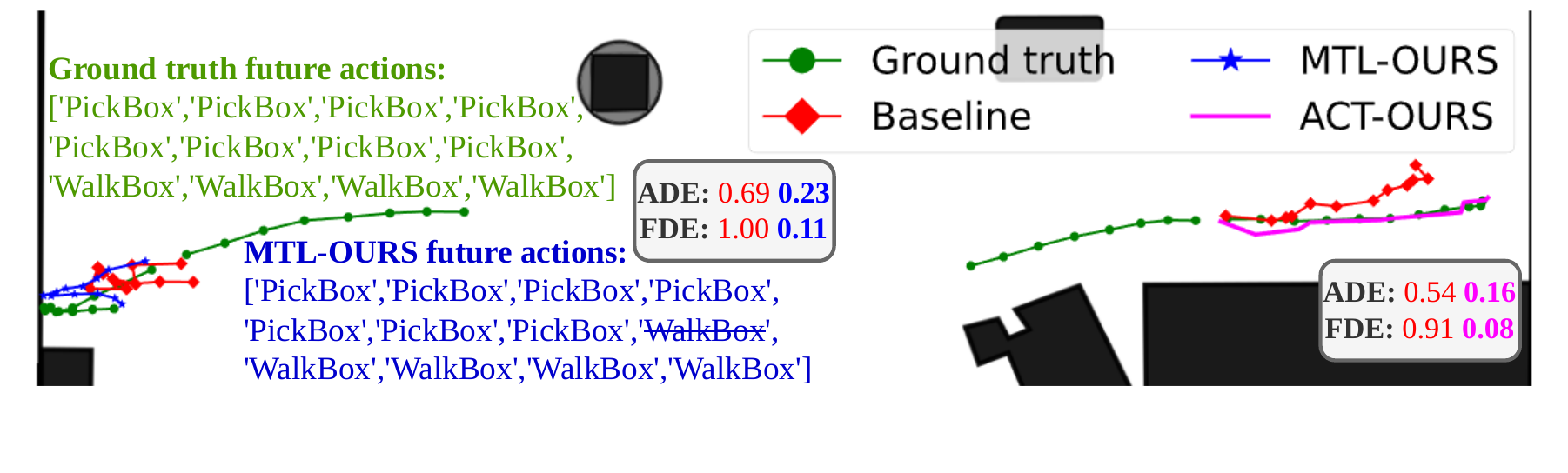}
  \vspace{-10mm}
  \caption{Prediction examples for \emph{Carrier--Box} in Scenario 3, for our multi-task learning framework (``MTL-OURS'', \textbf{left}) for joint trajectory and action prediction, and for our action-conditioned trajectory prediction (``ACT-OURS'', \textbf{right}), with a \SI{4.8}{s} prediction horizon.}
  \label{fig:pred}
  \vspace{-4mm}
\end{figure}

\textbf{Qualitative Results.}  \cref{fig:pred} highlights cases where action classes improve predictions.
For ``picking up a box" behavior, MTL reduces ADE/FDE errors from $0.69/1.00$ to $0.23/0.11$, with only one mispredicted future action. 
Similarly, action-conditioning reduces trajectory errors for ``dropping a box'' by leveraging observed action sequences.

\section{Conclusions}

The interplay between complex activities, actions, and locomotion dynamics of people and other agents still needs to be explored, particularly in industrial scenarios. 
This research gap can be addressed with comprehensive datasets that capture the relationship between actions and motion.
Our work introduces the TH\"{O}R-MAGNI~\emph{Act} dataset to align action labels with diverse human motion cues. 
These cues, including position data, head orientation, gaze, and semantic attributes, provide a rich description of human motion in industrial settings.
We also developed efficient and accurate transformer-based models for two applications where TH\"{O}R-MAGNI~\emph{Act} can play an important role: (1) action-conditioned trajectory prediction and (2) joint action and trajectory prediction. 
TH\"{O}R-MAGNI~\emph{Act} with the diverse annotation classes pave the way for future research in human motion modeling based on rich contextual cues.

\bibliographystyle{IEEEtran}
\bibliography{IEEEabrv,my_bib}

\end{document}